\documentclass[review]{elsarticle}
\usepackage{lineno,hyperref}
\modulolinenumbers[5]

\usepackage{graphicx}
\usepackage{amsmath}
\usepackage{amssymb}
\usepackage{subfigure}
\usepackage{stfloats}
\usepackage{bm}
\usepackage{bbm}
\usepackage{algorithm}
\usepackage{algorithmicx}
\usepackage{algpseudocode}
\usepackage{bm}
\usepackage{tabularx}
\usepackage{array}
\usepackage{multirow}
\usepackage{color}
\usepackage{url}
\usepackage{setspace}
\usepackage{mathrsfs}
\usepackage{flushend}

\usepackage{times}
\usepackage{epsfig}
\usepackage{makecell}
\usepackage{stfloats}
\usepackage{array}
\usepackage{color}
\usepackage{array}
\usepackage{booktabs} 
\usepackage{lscape}
\usepackage{bbding}

\journal{Journal of \LaTeX\ Templates}
\begin{document}

\begin{frontmatter}
\title{SelFLoc: Selective Feature Fusion for Large-scale Point Cloud-based Place Recognition}
\author[mymainaddress,mysecondaryaddress]{Qibo Qiu}
\ead{qiuqibo_zju@zju.edu.cn}
\author[mymainaddress]{Wenxiao Wang}
\author[mymainaddress]{Haochao Ying}
\author[mysecondaryaddress]{Dingkun Liang}
\author[mysecondaryaddress]{Haiming Gao\corref{mycorrespondingauthor}}
\cortext[mycorrespondingauthor]{Corresponding author}
\ead{ghm@mail.nankai.edu.cn}
\author[mymainaddress]{Xiaofei He}

\address[mymainaddress]{Zhejiang University}
\address[mysecondaryaddress]{Zhejiang Lab}

\begin{abstract}
Point cloud-based place recognition is crucial for mobile robots and autonomous vehicles, especially when the global positioning sensor is not accessible. LiDAR points are scattered on the surface of objects and buildings, which have strong shape priors along different axes. 
To enhance message passing along particular axes, Stacked Asymmetric Convolution Block (SACB) is designed, which is one of the main contributions in this paper. Comprehensive experiments demonstrate that asymmetric convolution and its corresponding strategies employed by SACB can contribute to the more effective representation of point cloud feature. On this basis, Selective Feature Fusion Block (SFFB), which is formed by stacking point- and channel-wise gating layers in a predefined sequence, is proposed to selectively boost salient local features in certain key regions, as well as to align the features before fusion phase.
SACBs and SFFBs are combined to construct a  robust and accurate architecture for point cloud-based place recognition, which is termed SelFLoc. Comparative experimental results show that SelFLoc achieves the state-of-the-art (SOTA) performance on the Oxford and other three in-house benchmarks with an improvement of $1.6$ absolute percentages on mean average recall@1. 
\end{abstract}


\end{frontmatter}


\section{Introduction}
\begin{figure}[t]
\begin{center}
	\includegraphics[width=12cm]{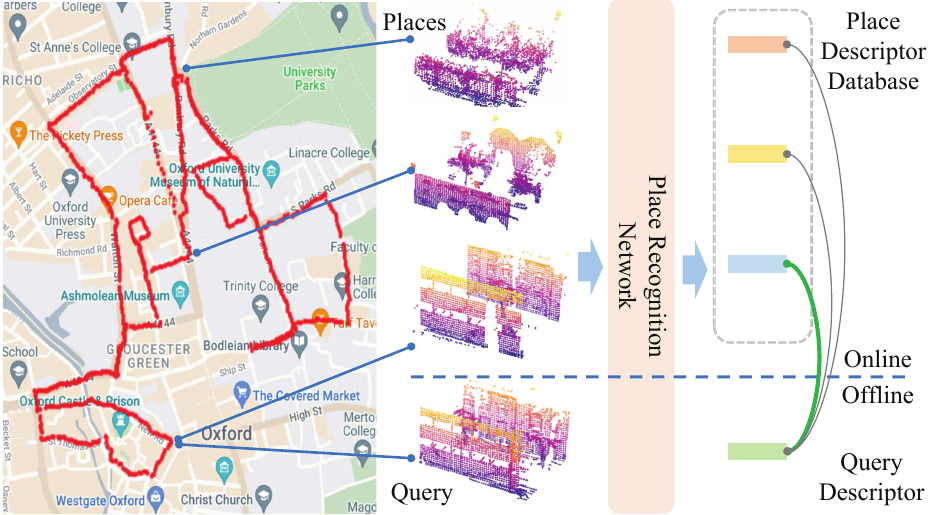}
\end{center}
\caption{Point cloud-based place recognition in large-scale urban environments. The place recognition network extracts global descriptors from point clouds in different locations, which are subsequently stored in a database. Once the vehicle reaches a new location, the closest match (\textcolor[rgb]{0.1, 0.65, 0.32}{\textbf{green}}) can be retrieved if the distance between the queried and recorded global descriptor is the shortest.}
\label{background}
\end{figure}

Due to the frequent absence of global positioning signals, localization based on environmental perception is becoming increasingly vital in the navigation system for both mobile robots and autonomous vehicles \cite{kbs_map, wang2022chase, kbs4}. Place recognition attempts to determine whether a place has been visited, which contributes to localization in GPS-denied environments. 
According to the type of sensor input, place recognition can be splited into image-based methods \cite{kbs2, pyriamid-vlad, kbs3} and point cloud-based methods. Point cloud-based methods are more robust to the lighting and seasonal changes \cite{pointnetvlad} when compared with image-based ones. 
In the remainder of this paper, \emph{place recognition} refers to LiDAR point cloud-based place recognition.
As shown in Fig. \ref{background}, the offline stage of place recognition collects the global descriptors of previously visited places, which forms the place descriptor database. The online stage extracts the global descriptor of current query LiDAR scan and ranks the LiDAR scans of previously visited places based on the Euclidean distances between the query descriptor and the descriptors in the database. 

With the development of deep learning, the representation of global descriptor for place recognition has progressed greatly. PointNetVLAD \cite{pointnetvlad} first creates the place recognition benchmarks and proposes a framework based on PointNet \cite{pointnet} and NetVLAD \cite{netvlad}. Different from images, point clouds are inherently unordered and irregular \cite{pointmixer}, which makes it difficult for PointNet-based models to capture the spatial relationships among the local neighboring points \cite{kbs5}. To this end, $k$-nearest neighbor ($k$-NN) \cite{lpd,ppt, ri_stv} and sparse voxelization-based 3D convolution \cite{minkloc3dv2,svt,transloc3d} are mostly utilized to capture the locality and hierarchy. Furthermore, transformer \cite{transformer, kbs6} is also employed for local feature enhancement \cite{ppt,soe,svt,transloc3d}. Nevertheless, aforementioned methods are proposed based on symmetrical sampling,
neglecting the fact that point clouds for various objects disperse differently along each axis.

In this paper, informative features are extracted by enhancing message passing along particular axes. Specifically, the Stacked Asymmetric Convolution Block (SACB) is designed, where multiple 1D asymmetric convolutions are stacked to extract local features, and the convolution numbers and strategies can vary with each axis to strengthen feature aggregation along particular axes. For example, additional convolution along $x$-axis can be performed to strengthen the feature in the driving direction, and dilation strategy can be further applied to expand the effective receptive field. In addition, the number of parameters can be reduced drastically compared with the traditional 3D convolution.  

On the other hand, to fuse multi-scale local features, Minkloc3D \cite{minkloc3d} is the first place recognition method that built on Feature Pyramid Network (FPN) to fuse multi-scale local features. The success of Minkloc3D-based models comes at fully use of the locality and hierarchy of 3D convolution, and shows that reliable place recognition is more dependent on local features. Salient local features locate at certain key regions, and feature semantics (channels) should be aligned before the fusion phase. Different from simple addition or concatenation, this paper argues that multi-scale features should be fused selectively, in which way local features can be boosted. Specifically, Selective Feature Fusion Block (SFFB) is introduced by stacking point- and channel-wise context gating layers to reweight the local features. Note that SFFB can be integrated as a plugin before any feature fusion phase.
The contributions of this paper are threefold:

\begin{itemize}
\item This paper proposes SACB to leverage the strong shape priors of the point cloud, which is stacked by 1D asymmetric convolutions equipped with different strategies. In addition, it reduces the parameters, which contributes to deployment.
\item SFFB block is introduced to fuse multi-scale features selectively, according to the point- and channel-wise context. Ablation experiments show that SFFB is beneficial to feature semantic alignment and key region enhancement, which further contributes to accurate global descriptor matching.
\item Comprehensive experiments show that both SACB and SFFB are effective for place recognition, supported by superior performance on the Oxford and three in-house datasets.
\end{itemize}

\section{Related Work}
\subsection{Point Cloud-based Place Recognition}
Traditional point cloud-based place recognition methods are depend on hand-crafted features \cite{m2dp,scancontext++,caoseason,luolidar}, which are well-designed to produce a discriminative global descriptor. Recently, the representation of discriminative feature is remarkably enhanced by deep learning methods. PointNetVLAD is the pioneering deep learning-based method proposed for place recognition, where the local features are extracted by PointNet. Different from images, it is difficult for PointNet-based models to capture the local response, since point clouds are inherently unordered and irregular \cite{pointmixer}. With the help of predefined local geometric features, LPD-Net \cite{lpd} can enhance local features by a graph-based aggregation operation. While NDT-Transformer \cite{ndt-transformer} transforms the point cloud into Normal Distribution Transform (NDT) cells, thus point-wise features can be boosted, and it is also the first one to make use of transformer for globality capturing. Furthermore, PPT-Net \cite{ppt} designs a pyramid point cloud transformer to capture globality spatially on different clustering granularities. 

Different from the aforementioned methods, 3D convolution can also be employed to extract local features. In particular, Minkloc3D \cite{minkloc3d} and its inherited versions are among the most successful 3D convolution-based place recognition models, which are built on the ResNet and FPN architectures. 
However, the proposed SelFLoc decomposes a 3D convolution into 1D convolutions to take advantage of the strong shape priors in point clouds, in addition to introducing novel attention mechanisms for selective feature fusion.
\subsection{Asymmetric Convolution}
Asymmetric convolution is originally designed to improve parameter efficiency. \cite{rethinking_inception} argues that any $n\times n$ kernel can be replaced by a 
$1\times n$ asymmetric convolution followed by a $n\times1$ convolution, and the computation cost can be greatly reduced. The hypothesis that a 2D kernel with a rank of one equals a sequence of 1D convolutions supports asymmetric convolution, while ranks of a 2D kernel cannot be guaranteed to be one. To this end, \cite{decomposeme} represents a 2D kernel (matrix) of rank $k$ as the outer product of a sequence of 1D convolutions (vectors). Supported by this low-rank approximation, the non-bottleneck module in ResNet \cite{resnet} is redesigned by ERFNet \cite{erfnet}, and this factorization considerably decreases the kernel size and enables real-time operation. ACNet \cite{acnet} provides a novel application of asymmetric convolution, which leverages 1D convolution to enhance the model robustness to rotational distortions. The SACB in proposed SelFLoc is not only intended for feature enhancement along particular axes and parameter reduction, but also equipping the asymmetric convolutions with different strategies.

\subsection{Attention and Context Gating Mechanisms}
Context gating as a reweighting operation designed to enhance the more informative features. There are both spatial and channel-wise attention mechanisms in image convolution networks \cite{ECAnet, SEnet, crn}. Likewise, point- and channel-wise attention mechanisms are both utilized in point feature aggregation \cite{kbs7}. In particular, PCAN \cite{pcan} presents a 3D point-wise attention map for place recognition and retrieval of point clouds, which is inspired by CRN \cite{crn}. On the other hand, the ECA \cite{ECAnet} module is introduced to place recognition by Minkloc3Dv2 \cite{minkloc3dv2} and TransLoc3D \cite{transloc3d}, where the global information is aggregated by channel-wise attention. 

Recently, transformer-based attention architecture has become popular in both natural language processing and computer vision. In addition, there are attempts to surpass the dominance of CNN and transformer by using MLP-like architecture \cite{mlp-mixer, pointmixer}. PointNet \cite{pointnet} is a pioneering trial in MLP-like architecture for point cloud. While PointMixer \cite{pointmixer} initially utilizes MLP-Mixer \cite{mlp-mixer} for point cloud understanding. In this paper, point- and channel-wise gating layers are employed for
semantic alignment and key region enhancement, which further contribute to reliable place recognition.

\section{Method}
\subsection{Asymmetric Convolution}
\textbf{Asymmetric convolution for point cloud. }
Initially, asymmetric convolution is created to reduce the computation cost and model size. ERFNet redesigns the non-bottleneck residual module of ResNet by introducing 1D asymmetric convolution, which contributes to traffic scene segmentation. In addition to reducing the computation cost, we replace the typical 3D convolution to decouple aggregation of features along each axis, which greatly enhances message passing along particular axes. Moreover, various strategies can be adapted for different 1D asymmetric convolution layers along different axes, e.g., dilation, deformation and stacking. 

A 3D convolution can be separated into a sequence of 1D convolutions, as supported by the low-rank hypothesis \cite{decomposeme}. Let $W \in \mathbb{R}^{d_{in} \times d_x \times d_y \times d_z \times d_{out}}$ denote the weights of a 3D convolution layer, where $d_{in}$ and $d_{out}$ are numbers of input and output planes, and $d_x \times d_y \times d_z$ indicates the kernel size. For convenience, $d_x$, $d_y$ and $d_z$ are set to the same value $d$. Given the $i$-th kernel in the 3D convolution layer $k^i \in \mathbb{R}^{d\times d \times d}$, it can be decomposed as follows:
\begin{equation}
	{k^{i}} = \sum_{r=1}^{R} \alpha_r^i \otimes \beta_r^i \otimes \gamma_r^i,
\end{equation}
where $R$ is the rank of $k^{i}$, $\otimes$ is the outer product operation, and $\alpha_r^i$, $\beta_r^i$, $\gamma_r^i$ are 1D vectors having the same size of $1 \times d$. In this paper, $R$ is set to $1$ for the trade-off between accuracy and efficiency. Therefore, the kernel $k^{i}$ with size of $d^3$ can be decomposed into three $1 \times d$ kernels, employing the decomposition on a 3D convolution layer with $3 \times 3 \times 3$ kernels can yield a $66\%$ reduction in parameters.

Buildings and objects are often aligned in the driving direction, and LiDAR points are typically distributed on their surfaces. What is more, point clouds scattered on various surfaces exhibit distinct correlations along different axes. For instance, assuming that the $x$-axis direction corresponds with vehicle orientation in a traffic scene, points on the lamp-post are distributed along the $z$-axis. Whereas points on the building wall are primarily scattered on the plane perpendicular to the $y$-axis. In general, correlation on the $x$- and $y$-axis is significant for points on the corners of buildings. 

\begin{figure}[t]
		\centering
		\subfigure[SACB]{\includegraphics[width=5.5cm]{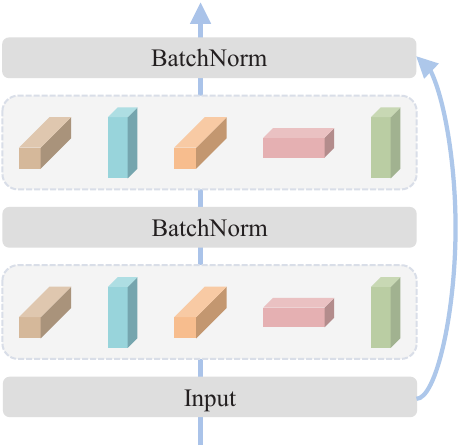}}
		\centering
		\subfigure[Asymmetric convolution]{\includegraphics[width=5.5cm]{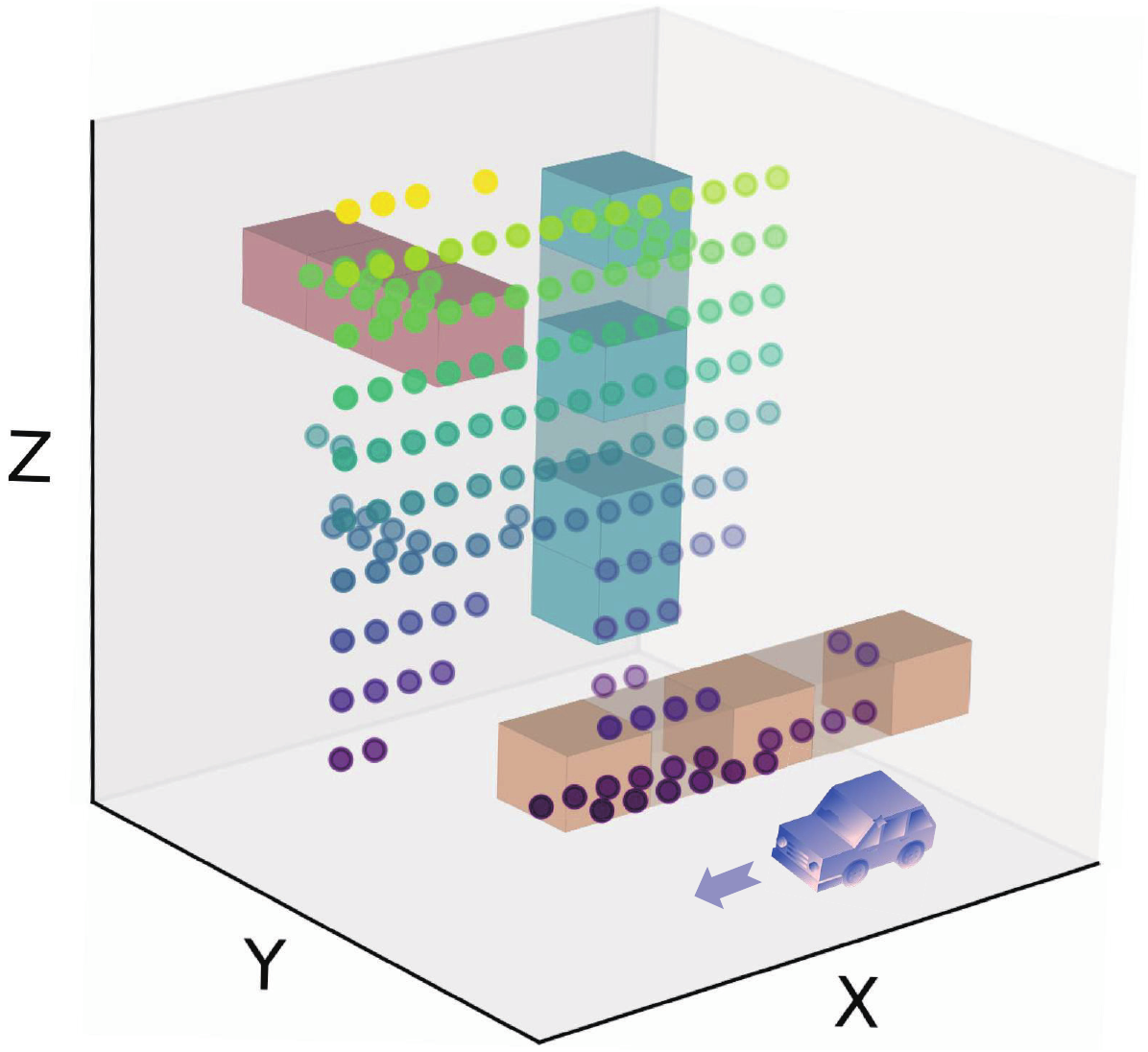}}
	\caption{(a): An SACB is composed of two sub-blocks, each of which is formed by stacking a specified number of asymmetric convolutions in a predefined sequence. (b): Asymmetric convolutions equipped with different strategies,  e.g., typical (\textcolor[rgb]{0.65 , 0.525, 0.525}{\textbf{pink}}), dilation (\textcolor[rgb]{0.713, 0.61, 0.528}{\textbf{orange}}) and deformation (\textcolor[rgb]{0.45, 0.636, 0.684}{\textbf{blue}}). }
	\label{SACB}
\end{figure}

To enhance the feature aggregation for a given scene, the ratio among the numbers of 1D convolutions along  $x$-, $y$- and $z$-axis can be customized to strengthen message passing along particular axes. As demonstrated in Fig. \ref{SACB} (a), each sub-block is created by stacking a series number of 1D convolutions along different axes in a predefined sequence. By the above decomposition, convolution operations along the $x$-, $y$- and $z$-axis can be conducted independently.  In addition, different strategies can be applied to these 1D convolutions along different axes, as illustrated in Fig. \ref{SACB} (b).

\subsection{Selective Feature Fusion}
In this paper, point- and channel-wise context gating layers are stacked to form the SFFB, which is a kind of mixed attention mechanism that contributes to local feature enhancement by leveraging two attention modes. SFFB works as a plugin placed before low- and high-level feature fusion, and the detailed design will be described as follows.

\textbf{Point-wise context gating.} It has been verified that point-wise context gating is beneficial to place recognition \cite{pcan}. Point clouds have plenty of low-level visual cues, e.g., edges, planes and corners, which are shaped by the corresponding key points. Moreover, certain regions with salient geometric shapes contribute the most to place recognition, e.g., doors, windows, lamp-posts and their spatial relationships.
We apply the point-wise attention to selectively boost the features of key regions. In addition, the operation contributes to key point estimation, which is crucial for the geometric verification task that follows the place recognition.

Given a feature map $X \in \mathbb{R}^{N \times C}$, $N$ and $C$ indicate the number of feature points and channels, respectively. A point-wise gating layer $F_{point}$ conducts a reweighting operation on each point:
\begin{equation}
	F_{point}(X_n)= X_n \cdot 
	\sigma\left(MLP_{point}\left(X_n\right)\right),
\end{equation}
where $X_n$ indicates the $n$-th point of $X$, $\sigma$ is the sigmoid activation. Given $X_n$ with size of $1 \times C$, the output size of $MLP_{point}$ is $1$. The point-wise layer has the same size of input and output, which pays more attention to key regions.
\begin{figure*}[ht]
	\centering
	\includegraphics[width=12cm]{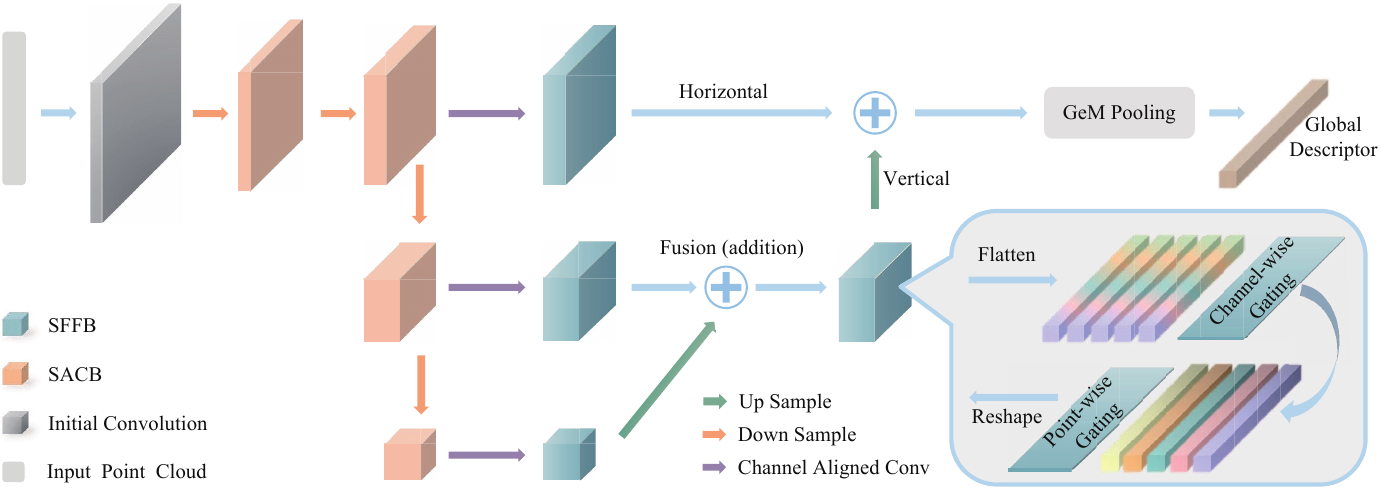}
	\caption{The architecture of SelFLoc implemented in an encoder-decoder style. In encoder stage, down sampling layers are utilized to reduce the resolutions of feature maps. each of which is followed by an SACB. Low- (horizontal) and high-level (vertical) features are fused (addition) during the decoder stage. Note that an SFFB is placed prior to the local feature fusion phase, which is intended for point- and channel-wise selective fusion refinement.}
	\label{selfloc_arch}
\end{figure*}

\textbf{Channel-wise gating.} A channel-wise gating layer conducts a reweighting operation on each channel, then channel dependencies can be exploited. Another concern is that place recognition usually aggregates local features into a global descriptor to compute the Euclidean distance. Channel-wise gating as a kind of semantic refinement further makes the semantics (channels) to be more comparable and more diverse, which contributes to accurate distance compution. The channel-wise gating layer $F_{channel}$ is represented as follows:
\begin{equation}
	F_{channel}(X_n)= X_n \odot  
	\sigma\left(MLP_{channel}\left(AvgPool\left(X\right)\right)\right),
\end{equation}
where $X_n \in \mathbb{R}^{1 \times C}$ indicates the $n$-th point of $X \in \mathbb{R}^{N \times C}$, $\odot $ and $\sigma$ are the Hadamard product (element-wise multiplication) and sigmoid activation, respectively. Notably, the value of $N$ is not fixed, due to the difference between the numbers of points in each frame, and an global average pooling operation is required to ensure the input size of $MLP_{channel}$ is $1 \times C$. 

Original SENet \cite{SEnet} uses global average pooling to squeeze global spatial information. Then fully-connected (FC) layers with dimensionality reduction are used to take advantage of the information. On the other hand, ECA reweights each channel by considering its nearest neighbors, thus only local cross-channel interaction is captured. Both of the above methods bring side effects on Euclidean distance computation. In this paper, FC layers without dimensionality reduction are employed in $MLP_{channel}$, which are more suitable for reliable place recognition. For a comprehensive comparison,  $MLP_{channel}$ in SENet, ECA-Net and MLP-Mixer styles are also implemented in our ablution experiments.

\subsection{Overall Architecture. }
As illustrated in Fig. \ref{selfloc_arch}, the overall architecture mainly consists of an initial convolution, SACBs and SFFBs, which is implemented in an encoder-decoder style. The encoder stage, in particular, adapts an initial convolution to project 3D LiDAR points into a deep feature space. Then multi-scale feature maps can be produced by a series of SACB blocks, placed after down sampling operations. Low- and high-level features are fused during the decoder stage. And the SFFB is naturally placed prior to the local feature fusion phase.

\textbf{Initial convolution} is conducted by a $K_0 \times K_0 \times K_0$ 3D sparse convolution for primary deep feature extraction. Given an input point cloud $P=\left \{ p_1,p_2,...,p_N \right \}$, initial feature ${X}=\left \{ {x_1}, {x_2},...,{x_N}\right \}$ can be extracted. $p_n$ indicates the coordinate of $n$-th point in the point cloud, whose size is $1 \times 3$, and the size of corresponding feature ${x_n}$ is $1 \times {C_0}$.
As described previously, the asymmetric convolution is more suitable for place recognition task than symmetric 3D convolution. However, asymmetric convolution is not guaranteed to work well on low-level layers \cite{rethinking_inception}, typical 3D sparse convolution is still required for this phase.

\textbf{SACB} consists of two sub-blocks, each sub-block has three asymmetric convolution layers operating on $x$-, $y$- and $z$-axis, respectively, as well as an additional layer along a particular axis. Furthermore, to verify the influence of additional convolution along different axes, three version are created:  \emph{SelFLoc\_X},  \emph{SelFLoc\_Y} and  \emph{SelFLoc\_Z}.
The dilation strategy is also introduced into the additional asymmetric convolution. For a fair comparison, only one SACB block of a specific depth employs dilation strategy for each trial.

Notice that the numbers and strategies of asymmetric convolutions along different axes can be varied and a study on these combinations may lead to superior outcome than the ones presented in this research. However, such studies are out of the scope of this research, and the above models are chosen for a good balance between accuracy and efficiency.

\textbf{SFFB} is placed prior to each additive fusion phase, which includes a point-wise context gating layer and a channel-wise one. There are two different stacking orders: point-wise gating first and channel-wise gating first. We implement these two model types, moreover, models with only point- or channel-wise gating layers are also implemented for comparison.

\textbf{Overall forward process} is elaborated in \textbf{Algorithm \ref{algo}}. The down and up sampling depths are denoted as $D_d$ and $D_u$, respectively, which should be predefined for a specific place recognition task. Note that the $ChannelAlignedConv$ is used for channel number alignment, which is inspired by Minkloc3dv2 and plays a crucial role for feature addition phase. In addition, GeM pooling \cite{GeM} is employed to aggregate local features into a global descriptor $G$ in accordance with typical place recognition methods. As discussed above, Euclidean distance employed in point cloud retrieval for place recognition demands the amplitudes of each channel are comparable. GeM, as a generalized form of harmonic and quadratic mean, can further align the channels (semantics).

\begin{algorithm}[!t]
\caption{The Overall Forward Process of SelFLoc.}
\begin{algorithmic}[1]
	\Require Point cloud $P$, down sampling depth $D_d$, up sampling depth $D_u$.
	\Ensure Global descriptor $G$ of point cloud $P$.
	\State Initialize lower level feature maps $F \leftarrow \emptyset$
	\State $X \leftarrow InitialConvolution(P)$
	\For{$d = 0 \to D_d -1$}
	\State $X \leftarrow DownSample(X)$
	\State $X \leftarrow SACB(X)$
	\If{$D_d - D_u -1 \le d < D_d-1$}
	\State $F \leftarrow F  \cup X$
	\EndIf
	\EndFor	
	\For{$d = 0 \to D_u -1$}
	\State $X \leftarrow UpSample(X)$
	\State $X \leftarrow SFFB(X)$
	\State Get the last $d$-th feature map $Y$ of $F$
	\State $Y \leftarrow ChannelAlignedConv(Y)$
	\State $Y \leftarrow SFFB(Y)$
	\State $X \leftarrow X+Y$
	\EndFor
	\State $G \leftarrow GeMPool(X)$
	\State Return $G$
\end{algorithmic}
\label{algo}
\end{algorithm}

\subsection{Probability Model of Point Cloud-based Place Recognition}
Given the training dataset $D=\{(q,i,j)\mid q \in\mathcal{Q}\}$,  where $\mathcal{Q}$ is the query set, and $i$ and $j$ indicate sampled point cloud frames. The training objective $\mathcal{O}$ is to maximize the posterior probability defined as follows:
\begin{align}
	\mathcal{O} & =\ln p\left(\Theta \mid D \right) \nonumber \\
	& =\ln p\left(D \mid \Theta\right) p(\Theta) \label{eq5},
\end{align}
where $\Theta$ indicates the parameters of SelFLoc. According to reference~\cite{bpr}, we introduce a common assumption that $p(\Theta)$ is a normal distribution: $p(\Theta)\sim N(0,\lambda_{\Theta}I)$. $p\left(D \mid \Theta\right)$ is the likelihood function, which can be rewriten as follows:
\begin{equation}
	 \ln p\left(D \mid \Theta\right) = \ln \prod_{(q, i, j) \in D} p\left(i>_{q} j \mid \Theta\right)^{\delta((q,i,j){\in}D_O)}\cdot (1-p\left(i>_{q} j \mid \Theta\right))^{\delta((q,i,j){\notin}D_O)},
\end{equation}
where $\delta$ is the indicator function. $p\left(i>_{q} j \mid \Theta\right)$ represents the probability that the point cloud frame $i$ is closer to the query frame $q$ compared to the frame $j$. $D_O$ is a set containing observed preferences, which can be defined as: 
\begin{equation}
D_O=\{(q,i,j)\mid q \in\mathcal{Q} \wedge i \in S_P \wedge j \in S_N \},
\end{equation}
where $S_P$ and $S_N$ represent the positive and negative sets, respectively. Moreover, the probability can be calculated by the corresponding global descriptors as:
\begin{equation}
	p(i>_cj|\Theta)=\sigma(\hat{G}_{qij}(\Theta))
\end{equation}
\begin{equation}
	\hat{G}_{qij}(\Theta) = d(q,i) - d(q,j) 
\end{equation}
\begin{equation}\sigma(x):=\frac{1}{1+e^{-x}},\end{equation}
where $d(q,i)$ is the Euclidean distance between the global descriptors of $q$ and $i$. Combining the above equations, we rewrite the objective function as follows:
\begin{align}
	\mathcal{O} &= \ln \prod_{(q, i, j) \in D} p\left(i>_{q} j \mid \Theta\right)p(\Theta) \nonumber \\
	&= \ln \prod_{(q, i, j) \in D} \sigma(\hat{G}_{qij}(\Theta))^{\delta((q,i,j){\in}D_O)}(1-\sigma(\hat{G}_{qij}(\Theta)))^{\delta((q,i,j){\notin}D_O)} p(\Theta) \nonumber \\
	&=\sum_{(q,i,j)\in D_O}\ln\sigma(\hat{G}_{qij}(\Theta))+\sum_{(q,i,j)\notin D_O}\ln(1-\sigma(\hat{G}_{qij}(\Theta)))+\ln p(\Theta)\nonumber \\
	&=\sum_{(q,i,j)\in D_O}\ln\sigma(\hat{G}_{qij}(\Theta))+\sum_{(q,i,j)\notin D_O}\ln(1-\sigma(\hat{G}_{qij}(\Theta)))-\lambda_\Theta||\Theta||^2 \nonumber \\
	&\approx \sum_{(q,i,j)\in D_O}\ln\sigma(d(q,i) - d(q,j))+\sum_{(q,i,j)\notin D_O}\ln(1-\sigma(d(q,i) - d(q,j))).
\end{align}
According to Jensen's inequality, the objective function has a lower bound:
\begin{align}
	\mathcal{O} &\ge \ln \sum_{(q,i,j)\in D_O}\sigma(d(q,i) - d(q,j)) + \ln \sum_{(q,i,j)\notin D_O}(1-\sigma(d(q,i) - d(q,j))) \nonumber \\ 
	&= \ln \mathcal{O}_{1} + \ln (|D - D_O| - \mathcal{O}_2) \label{lb}.
\end{align}
According to the monotonicity of Equation~\ref{lb}, maximizing the lower bound is equivalent to maximizing $\mathcal{O}_{1}$ and minimizing $\mathcal{O}_{2}$. Training the proposed SelFLoc using the Smooth-AP~\cite{smoothap,minkloc3dv2} loss function can achieve this goal.
\\
\textbf{Relation to Smooth-AP Loss.} Smooth-AP is a commonly used training loss function in the field of place recognition. Specifically, it calculates $AP_q$ for each query frame $q$ as follows:
\begin{align}
	AP_{q}&\approx\frac{1}{|S_{P}|}\sum_{i\in{S}_{P}}\frac{1+\sum_{h\in{S}_{P}}\sigma(d(q,i) - d(q,h))}{1+\sum_{h\in{S}_{P}}\sigma(d(q,i) - d(q,h))+\sum_{j\in{S}_{N}}\sigma(d(q,i) - d(q,j))} \nonumber \\
	& =  \frac{1}{|{S}_{P}|}\sum_{i\in{S}_{P}}\frac{1+ \mathcal{O}_{2}}{1+\mathcal{O}_{2}+ \mathcal{O}_{1}}.\label{ap}
\end{align}
Therefore, the loss for each batch can be calculated as:
\begin{align}
	\mathcal{L}_{AP} &=\frac{1}{m}\sum_{q=1}^{m}{(1-AP_q)},\label{loss}
\end{align}
where $m$ is the number of queries in one batch. Comparing Equation~\ref{lb}, \ref{ap}, and \ref{loss}, we can see that maximizing $\mathcal{L}_{AP}$ is equivalent to maximizing the lower bound of objective function $\mathcal{O}$.

\section{Experiments}
This section verifies the performance of proposed SelFLoc by conducting experiments on a variety of benchmark datasets. Additionally, ablation studies are performed to analyze the introduced blocks and strategies.
\subsection{Experiments Setting}
\textbf{Benchmark and evaluation. } PointNetVLAD originally created four benchmark datasets for evaluating point cloud-based place recognition networks: Oxford is a partial set of Oxford RobotCar dataset \cite{Oxford}, U.S., R.A. and B.D. are respective in-house datasets of a university sector, a residential area and a business district. These datasets are obtained using a LiDAR sensor installed on a car that regularly drives over each of the four regions at different times, and collects data under varying environmental conditions.

The locations are sampled at a specified interval (shown in Table \ref{datasets}) along continuous tracks of the vehicle, and the corresponding submaps are constructed by dividing the LiDAR scans and erasing non-informative ground planes. Each submap is down sampled to 4096 points using a voxel filter and tagged with the Universal Transverse Mercator (UTM) coordinate of its centroid. During training tuple generation, a point cloud pair is considered positive if the distance is less than 10m and negative if the distance is greater than 50m. In order to evaluate various place recognition methods, the query point cloud is successfully localized when the retrieved point cloud is within 25m. 

Our evaluation follows the baseline training pipeline introduced in \cite{pointnetvlad}. Specifically, the network is trained using the training set of the Oxford dataset and tested on the testing sets of the Oxford and three in-house datasets. Table \ref{datasets} shows the details of training and testing datasets, the average recall@1\% (AR@1\%) and average recall@1 (AR@1) metrics are primarily adopted for evaluation.

\begin{table}[]
	\centering
	\caption{The number of submaps for training and testing, * approximate value.}
	\resizebox{0.7\linewidth}{!}{
	\begin{tabular}{cccccc}
		\hline \hline
		\specialrule{0em}{1pt}{1pt}
		& \makecell{Training \\Set}          & \makecell{Testing\\ Set}             & \makecell{Submaps\\/Run} & \makecell{Intervals\\(training)} & \makecell{Intervals\\(testing)}\\ 
		\hline
		\specialrule{0em}{1pt}{1pt}
		Oxford & 21711                 & 3030                  & 120-150    & 10m &20m \\ \hline
		\specialrule{0em}{1pt}{1pt}
		U.S.   & \multirow{3}{*}{6671} & \multirow{3}{*}{4542} & 400*   & \multirow{3}{*}{12.5m} & \multirow{3}{*}{25m}     \\ 
		R.A.   &                       &                       & 320*        \\ 
		B.D.   &                       &                       & 200*        \\ \hline \hline
	\end{tabular}
}
	\label{datasets}

\end{table} 

\textbf{Implementation details. }Implementation of the proposed network SelFLoc is based on the MinkLoc3Dv2 codebase, where the sparse convolution on point clouds is implemented by Minkowski Engine \cite{ME}.

To be fair, the size of input point clouds is $4096 \times 3$, which is the same as PointNetVLAD. Each down sampling layer decreases the spatial resolution by two. In our experiments, there are $4$ down sampling layers (each is placed before an SACB), and $2$ up sampling layers (each is followed by an SFFB). The training loss, stays the same with MinkLoc3Dv2. 

\begin{figure*}[!ht]
	\begin{center}
		\includegraphics[width=12cm]{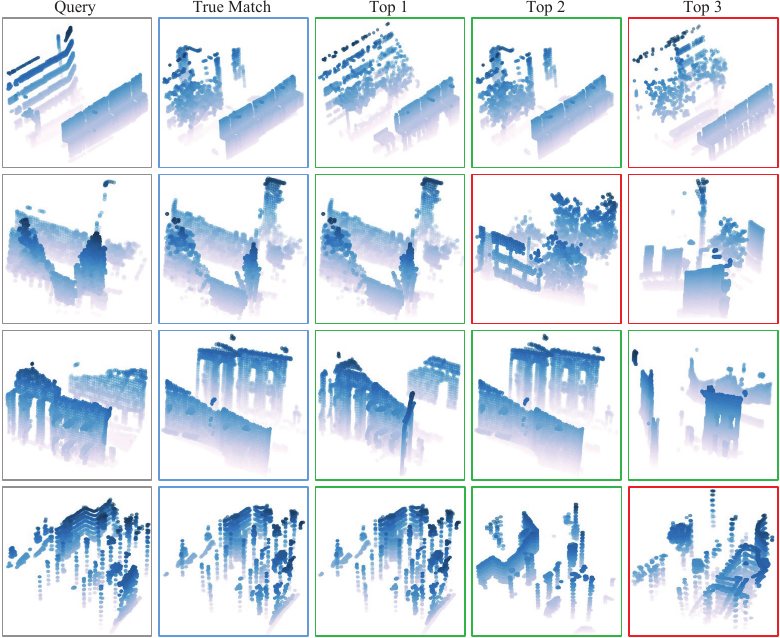}
	\end{center}
	\caption{Query (gray) and top $3$ retrieved frames (green: successful, red: failed). Moreover, one of the true (blue) matches is displayed for comparison. SelFLoc successfully finds the closest match even when the perspective changes (row 3). }
	\label{demos}
\end{figure*}
\subsection{Main Results}
To verify the quantitative performance of the proposed SelFLoc method, we conduct comprehensive experiments on the benchmark datasets in \cite{pointnetvlad}. Specifically, we compare SelFLoc with a lot of advanced methods, including RI\_STV \cite{ri_stv}, MinkLoc3Dv2 \cite{minkloc3dv2}, HiBi-Net \cite{shu2023hierarchical}, SVT-Net \cite{svt}, PPT-Net \cite{ppt}, NDT-transformer \cite{ndt-transformer}, EPC-Net \cite{epc-net}, LPD-Net \cite{lpd}, PCAN \cite{pcan} and the pioneering PointNetVLAD \cite{pointnetvlad}. Table \ref{main_res_base} reports the AR@1\% and AR@1 metrics of aforementioned methods trained by the baseline pipeline. Although many excellent methods have be proposed for place recognition, and there is little room for improvement \cite{minkloc3dv2, transloc3d}. SelFLoc can still has a remarkable performance, outperforming the most sophisticated methods \cite{ri_stv, minkloc3dv2} by $1.6$ absolute percentages on AR@1 (90 vs. 91.6) and $1$ absolute percentage on AR@1\% (95.1 vs. 96.1).
Moreover, the in-house datasets in Table \ref{main_res_base} have not be trained and the AR@1 metric has been improved by 2.3\%, 2.3\% and 2.1\%, respectively. Compared with MinkLoc3Dv2, SelFLoc has a better capability for generalization in addition to its high accuracy. This is crucial for mobile robots because their operation scenarios are diverse, and it is difficult to collect sufficient training data for each scenario.
Note that 3D convolution-based methods have higher metrics than $k$NN-based and NDT-based methods in the mass, which demonstrates that voxel-based convolution is still an effective technique to capture the locality and hierarchy of point cloud. 

\begin{table}[!h]
	\centering
	\caption{Evaluation results of the advanced place recognition methods.}
	\resizebox{\linewidth}{!}{
	\begin{tabular}{ccccccccccc}
		\hline \hline
		\specialrule{0em}{1pt}{1pt}
		& \multicolumn{2}{c}{Oxford} & \multicolumn{2}{c}{U.S.}& \multicolumn{2}{c}{R.A.}&\multicolumn{2}{c}{B.D.}& \multicolumn{2}{c}{Mean} \\ 
		\multirow{-2}{*}{} & \multicolumn{1}{c}{AR@1} & {AR@1\%} & \multicolumn{1}{c}{AR@1} & {AR@1\%} & \multicolumn{1}{c}{AR@1} & {AR@1\%} & \multicolumn{1}{c}{AR@1} & {AR@1\%} & \multicolumn{1}{c}{AR@1} & {AR@1\%} \\ \hline
		\specialrule{0em}{1pt}{1pt}
		 PointNetVLAD~\cite{pointnetvlad}       & \multicolumn{1}{c}{62.8}                              & 80.3                              & \multicolumn{1}{c}{63.2}  & 72.6                              & \multicolumn{1}{c}{56.1}  & 60.3                              & \multicolumn{1}{c}{57.2}  & 65.3                              & \multicolumn{1}{c}{59.8}  & 69.6                              \\ 
		 PCAN~\cite{pcan}               & \multicolumn{1}{c}{69.1}                              & 83.8                              & \multicolumn{1}{c}{62.4}  & 79.1                              & \multicolumn{1}{c}{56.9}  & 71.2                              & \multicolumn{1}{c}{58.1}  & 66.8                              & \multicolumn{1}{c}{61.6}  & 75.2                              \\ 
		LPD-Net~\cite{lpd}            & \multicolumn{1}{c}{86.3}                              & 94.9                              & \multicolumn{1}{c}{87.0}  & 96.0                              & \multicolumn{1}{c}{83.1}  & 90.5                              & \multicolumn{1}{c}{82.5}  & 89.1                              & \multicolumn{1}{c}{84.7}  & 92.6                              \\ 
		EPC-Net~\cite{epc-net}            & \multicolumn{1}{c}{86.2}                              & 94.7                              & \multicolumn{1}{c}{-}     & 96.5                              & \multicolumn{1}{c}{-}     & 88.6                              & \multicolumn{1}{c}{-}     & 84.9                              & \multicolumn{1}{c}{-}   & 91.2                              \\ 
		SOE-Net~\cite{soe}          & \multicolumn{1}{c}{-}                                 & 96.4                              & 
		\multicolumn{1}{c}{-}     & 93.2                              & \multicolumn{1}{c}{-}     & 91.5                              & \multicolumn{1}{c}{-}     & 88.5                              & \multicolumn{1}{c}{-}     & 92.4                              \\ 
		HiBi-Net~\cite{shu2023hierarchical}            & \multicolumn{1}{c}{87.5}                                 & 95.1                              & 
		\multicolumn{1}{c}{87.8}     & -                              & \multicolumn{1}{c}{85.8}     & -                              & \multicolumn{1}{c}{83.0}     & -                              & \multicolumn{1}{c}{86.0}     & -                              \\ 
		MinkLoc3D~\cite{minkloc3d}          & \multicolumn{1}{c}{93.0}                              & 97.9                              & \multicolumn{1}{c}{86.7}  & 95.0                              & \multicolumn{1}{c}{80.4}  & 91.2                              & \multicolumn{1}{c}{81.5}  & 88.5                              & \multicolumn{1}{c}{85.4}  & 93.2                              \\ 
		NDT-Transformer~\cite{ndt-transformer}    & \multicolumn{1}{c}{93.8}                              & 97.7                              & \multicolumn{1}{c}{-}     & -                                 & \multicolumn{1}{c}{-}     & -                                 & \multicolumn{1}{c}{-}     & -                                 & \multicolumn{1}{c}{-}   & -                                 \\ 
		PPT-Net~\cite{ppt}            & \multicolumn{1}{c}{93.5}                              & 98.1                              & \multicolumn{1}{c}{90.1}  & 97.5                              & \multicolumn{1}{c}{84.1}  & 93.3                              & \multicolumn{1}{c}{84.6}  & 90.0                              & \multicolumn{1}{c}{88.1}  & 94.7                              \\ 
		SVT-Net~\cite{svt}            & \multicolumn{1}{c}{93.7}                              & 97.8                              & \multicolumn{1}{c}{90.1}  & 96.5                              & \multicolumn{1}{c}{84.3}  & 92.7                              & \multicolumn{1}{c}{85.5}  & 90.7                              & \multicolumn{1}{c}{88.4}  & 94.4                              \\ 
		TransLoc3D~\cite{transloc3d}         & \multicolumn{1}{c}{95.0}                              & 98.5                              & \multicolumn{1}{c}{-}     & 94.9                              & \multicolumn{1}{c}{-}     & 91.5                              & \multicolumn{1}{c}{-}     & 88.4                              & \multicolumn{1}{c}{-}   & 93.3                              \\ 
		MinkLoc3Dv2~\cite{minkloc3dv2}        & \multicolumn{1}{c}{\textbf{96.3}}                              & \textbf{98.9}                              & \multicolumn{1}{c}{90.9}  & 96.7                              & \multicolumn{1}{c}{86.5}  & 93.8                              & \multicolumn{1}{c}{86.3}  & 91.2                              & \multicolumn{1}{c}{90}    & 95.1                              \\ 
		RI\_STV~\cite{ri_stv}       & \multicolumn{1}{c}{-}                              & 98.5                              & \multicolumn{1}{c}{-}  & 97.3                              & \multicolumn{1}{c}{-}  & 93.0                              & \multicolumn{1}{c}{-}  & 91.7                             & \multicolumn{1}{c}{-}    & 95.1                              \\ 
		SelFLoc (ours)     & \multicolumn{1}{c}{96.0}                             & 98.8                             & \multicolumn{1}{c}{\textbf{93.2}} & \textbf{98.3}                             & \multicolumn{1}{c}{\textbf{88.8}} & \textbf{94.8}                             & \multicolumn{1}{c}{\textbf{88.4}} & \textbf{92.4}                             & \multicolumn{1}{c}{\textbf{91.6}} & \textbf{96.1}                             \\ \hline \hline
	\end{tabular}
}
	\label{main_res_base}
\end{table}

\begin{figure}[!]
	\begin{center}
		\includegraphics[width=12.5cm]{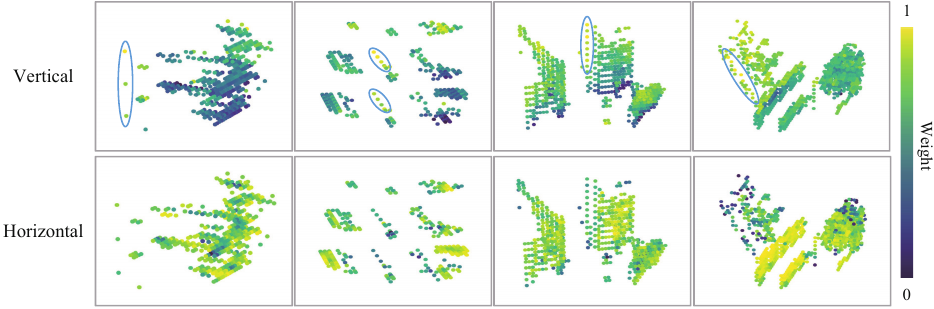}
	\end{center}
	\caption{Point-wise attention in horizontal and vertical SFFBs. Note that the horizontal point-wise attention is generated by lower-level local features, while the vertical one is selectively enhanced by semantics from higher level.  Points in blue circles are isolated but salient.}
	\label{attn_point}
\end{figure}

Fig. \ref{demos} displays the query point cloud and the point clouds retrieved by SelFLoc. The cases are challenging and SelFLoc successfully retrieves the closest match by leveraging the most discriminative feature. As depicted in Fig. \ref{selfloc_arch}, SFFB can be plugged into both horizontal and vertical feature branches,
we further visualize the point-wise attention maps in these branches before the last fusion phase.
As shown in Fig. \ref{attn_point}, point-wise gating layer in horizontal SFFB mainly pays attention to the major structure of the point cloud, while point-wise attention in vertical branch can be selectively allocated to the points isolated but salient.

\subsection{Ablation Study}
\begin{figure*}[]
	\begin{center}
		\includegraphics[width=12cm]{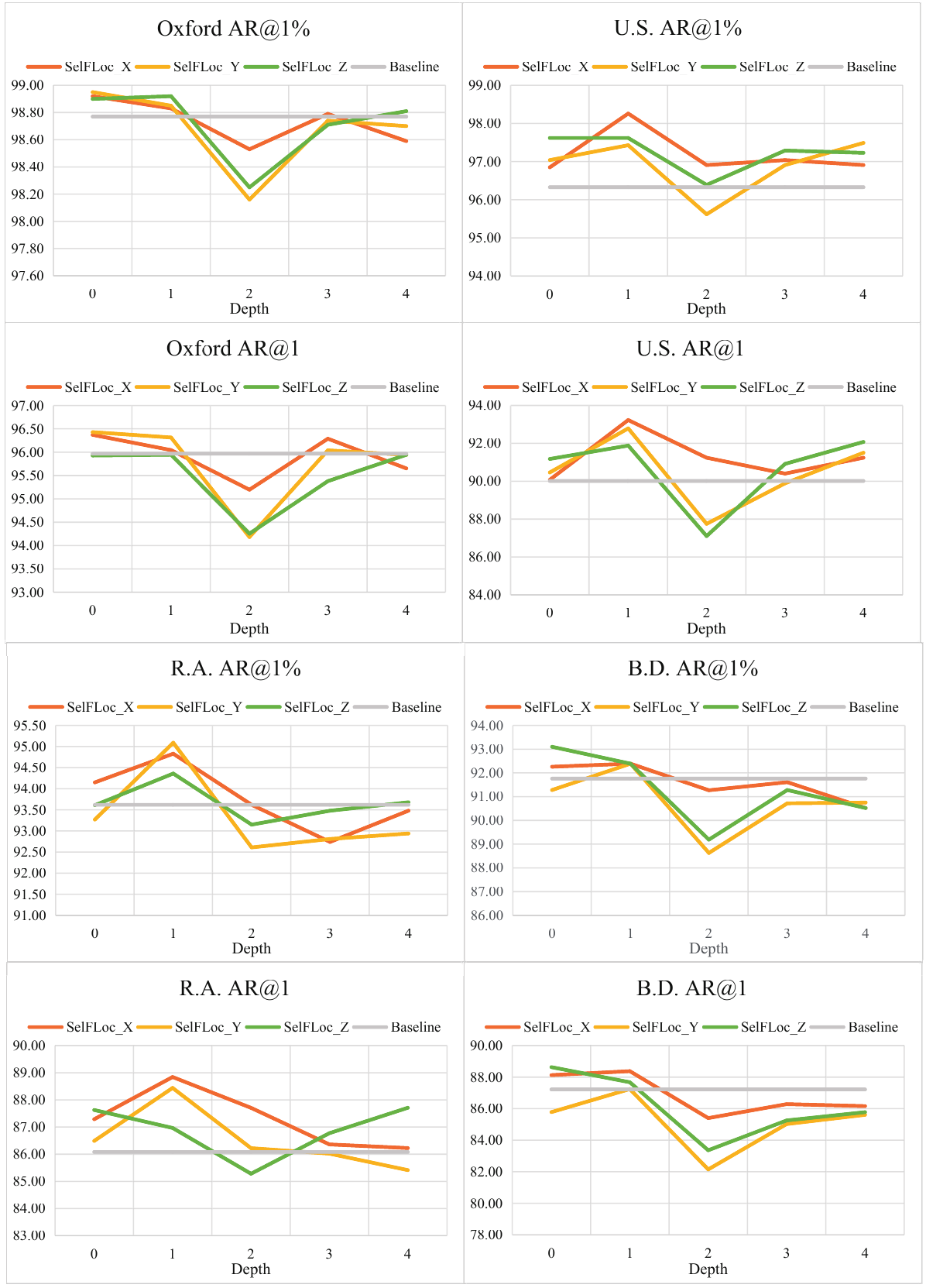}
	\end{center}
	\caption{Horizontal axis represents the depth of which SACB equipped with dilation strategy. There are $4$ SACBs employed in our experiments, and $Depth=0$ indicates that no SACB is equipped with dilation strategy. \emph{SelFLoc\_X}, \emph{SelFLoc\_Y} and \emph{SelFLoc\_Z} represents the models with one additional layer along $x$-, $y$- and $z$-axis, respectively, while model without additional asymmetric convolution layer is regarded as the \textbf{baseline}. Note that the dilation strategy is only applied on the additional layer of each sub-block in the SACB.}
	\label{dil_add}
\end{figure*}

\textbf{Additional axes. }To verify the impact of different additional layers, asymmetric convolution layers along $x$-, $y$- and $z$-axis are respectively added to the sub-blocks of SACBs every time. Fig. \ref{dil_add} displays the quantitative results with different additional layers, and \emph{SelFLoc\_X} achieves better accuracy and robustness, demonstrating that feature extraction along the $x$-axis is the most effective for the place recognition task. This may inspire researchers in the field of mobile robotics to optimize axis-oriented point cloud processing methods for specific scenarios. For example, in the field of self-driving, there may be more sophisticated methods developed in the future for enhancing the transmission of point cloud features along X-axis.

\textbf{Dilation strategy. }The effective receptive field can be expanded by utilizing the dilation strategy. Results of testing dilation influence on different depths of SACBs in Fig. \ref{dil_add} indicate that models with lower level dilated convolutions have higher metrics. Dilated convolutions in lower layers contribute more to the capture of local response. On the contrary, dilated convolutions in higher layers help to capture the interrelationship between semantics at higher levels, however the performance will be negatively affected if levels of dilated convolutions are excessively high. The performance of dilated convolution layers in different depths evidences that place recognition relies more on local features, and applicable depth is crucial for dilation strategy, e.g., $Depth=1$.

\textbf{Channel-wise attention mechanisms. }In order to compare the effectiveness of different attention mechanisms for channel-wise feature aggregation, we conduct experiments with various attention mechanisms. Specifically, FC layers without dimensionality reduction (SelFLoc-FC), channel-wise gating in the SENet (SelFLoc-SE), channel-wise gating in the ECA-Net (SelFLoc-ECA) and channel-wise attention proposed in MLP-Mixer (SelFLoc-Mixer) are implemented, as shown in Table \ref{channel_attn}. The comparison between SelFLoc-Mixer and other methods shows that  the squeeze phase is crucial for both robustness and accuracy. Moreover, FC attention without dimensionality reduction or neighbor limitation further improves the robustness by a large margin.
\begin{table}[!h]
	\centering
	\caption{Evaluation results (AR@1) of models with different channel-wise attention mechanisms.}
	\vspace{5pt}
	\resizebox{0.6\linewidth}{!}{
	\begin{tabular}{ccccc}
		\hline \hline
		\specialrule{0em}{1pt}{1pt}
		Model Type    & Oxford         & U.S.           & R.A            & B.D.           \\  \hline
		\specialrule{0em}{1pt}{1pt}
		SelFLoc-ECA            &  \textbf{96.53}          & 90.72          & 87.37          & 87.35          \\ 
		SelFLoc-SE             & 96.39          & 90.47          & 87.50          & 87.05          \\ 
		SelFLoc-Mixer          & 94.59              & 88.73             & 86.30             & 84.47  \\ 
		SelFLoc-FC(Ours)           &96.04 & \textbf{93.23} & \textbf{88.84} & \textbf{88.38} \\ \hline \hline
	\end{tabular}
}
\label{channel_attn}
\end{table}

\begin{table}[!h]
\centering
\caption{Evaluation results (AR@1) of different gating orders.}
\resizebox{0.6\linewidth}{!}{
\begin{tabular}{cccccc}
	\hline \hline
	\makecell{First\\Layer} & \makecell{Second\\Layer} & Oxford         & U.S.           & R.A            & B.D.           \\  \hline
	P           & P            & 96.11          & 92.33          & 85.89          & 86.84          \\ 
	P           & C            & \textbf{96.21} & 91.76          & 86.76          & 86.31          \\ 
	C           & P            & 96.04          & \textbf{93.23} & \textbf{88.84} & \textbf{88.38} \\ 
	C           & C            & 94.95          & 90.85          & 88.17          & 87.19          \\ \hline \hline
\end{tabular}
}
\label{cp_order}
\end{table}
\begin{table}[!h]
\centering
\caption{Ablation study results (AR@1) of different D1 and D2.}
\resizebox{0.6\linewidth}{!}{
\begin{tabular}{cccccc}
\hline \hline
\specialrule{0em}{0.5pt}{0.5pt}
D1 & D2 & Oxford         & U.S.           & R.A            & B.D.  \\ \hline 
\specialrule{0em}{0.5pt}{0.5pt}
128 & 128 & 94.88          & 87.75          & 86.76          & 83.92          \\ 
256 & 128 & 95.29          & 90.47          & 83.94          & 84.57          \\ 
256 & 256 & 96.04          & 93.23 & \textbf{88.84} & \textbf{88.38} \\ 
256 & 512 & \textbf{96.24} & 92.65          & 87.83          & 86.72          \\ 
512 & 512 & 96.10          & \textbf{93.43}          & 88.44          & 87.90          \\ \hline \hline
\end{tabular}
}
\label{dim}
\end{table}
\textbf{Combinations of point- and channel-wise gating layers. } An SFFB is composed of a point-wise and a channel-wise gating layer. Table \ref{cp_order} shows performance of different combinations of 
gating layers. From the table, it can be seen that SFFBs with only point or channel-wise gating layer have worse performance than SFFBs with both point- and channel-wise gating layers. In addition,  the best performance can be obtained when the first layer is channel-wise and the second layer is point-wise. As mentioned previously, channel-wise gating is employed for semantic (channel) alignment, therefore earlier alignment can result in better performance.

\textbf{Dimension influence. } Here we study the impact of model size. In particular,
we denote D1, D2 as the output dimensions of last and penultimate SACB blocks, respectively. Note that D2 is equal to the numbers of channels in each fusion phase, as well as the size of global descriptor for retrieval. As shown in Table \ref{dim}, expanding the model size can lead to higher accuracy, while overfitting will reduce its robustness. The model with $D1 = 256$ and $D2 = 256$ achieves an empirically optimal trade-off between accuracy and robustness.
\section{Conclusion}
In this paper, a novel architecture named SelFLoc is proposed for point cloud-based place recognition, which takes advantage of the strong shape priors of point clouds by stacking 1D asymmetric convolutions equipped with different strategies. In addition, features from different scales are refined by selective point- and channel-wise gating layers before the fusion phase. Comprehensive experiments on the Oxford dataset and three in-house datasets demonstrate that SelFLoc can achieve SOTA performance in terms of both accuracy and robustness. 
\\
\textbf{Limitation and Future Work.} The decomposition of 3D convolution, enhancement of axis-oriented features, and selective feature fusion in this research are all tailored for scenarios involving self-driving. However, the introduction of these strategies relies on the experience of researchers. To make them more applicable to other robotic studies, such as drones and bipedal robots, we will explore novel methods to endow robots with the ability to autonomously learn these strategies. One potential approach is to integrate the selection of strategies into the learnable world model~\cite{daydreamer,end-to-end}.
\section{Acknowledgements}
This research is supported in part by the National
Natural Science Foundation of China under Grant 62303428, and in part by Zhejiang Provincial Natural Science Foundation of China under Grant LQ23F030010.
\bibliography{KNOSYS_111794}

\end{document}